\RequirePackage{fix-cm}
\documentclass[twocolumn]{svjour3}          
\smartqed  
\usepackage{graphicx}
\usepackage[ruled,vlined]{algorithm2e}
\usepackage{array,tabularx}
\usepackage{xcolor}
\usepackage{eucal}
\usepackage{multirow}
\usepackage{amsmath,amssymb}
\usepackage{cite}
\usepackage{booktabs} 
\usepackage{microtype}

\usepackage{hyperref}
\hypersetup{
    colorlinks=true,
    linkcolor=green, 
    filecolor=teal, 
    urlcolor=purple, 
    pdftitle={Overleaf Example},
    pdfpagemode=FullScreen,
    }

\newenvironment{conditions*}
  {\par\vspace{\abovedisplayskip}\noindent
  \tabularx{\columnwidth}{>{$}l<{$} @{}>{${}}c<{{}$}@{} >{\raggedright\arraybackslash}X}}
  {\endtabularx\par\vspace{\belowdisplayskip}}

\newcommand{\revised}[1]{\textcolor{black}{{#1}}}

\DeclareMathOperator{\E}{\mathbb{E}}
\DeclareMathOperator*{\argmax}{arg\,max}  

\begin{document}

\title{Deep Reinforcement Learning for Multi-Agent Coordination}
\titlerunning{Deep RL for Multi-agent Coordination} 

\author{Kehinde O. Aina         \and
        Sehoon Ha }
\authorrunning{K.O. Aina and S. Ha} 

\institute{Kehinde O. Aina \at
              Institute for Robotics and Intelligent Machines, Georgia Institute of Technology, Atlanta Georgia, USA \\
              \email{kaina3@gatech.edu}           
           \and
           Sehoon Ha \at
              Institute for Robotics and Intelligent Machines, Georgia Institute of Technology, Atlanta Georgia, USA.\\
              \email{sehoonha@gatech.edu}    
}

\date{Received: date / Accepted: date}

\maketitle

\begin{abstract}

We address the challenge of coordinating multiple robots in narrow and confined environments, where congestion and interference often hinder collective task performance.
Drawing inspiration from insect colonies, which achieve robust coordination through stigmergy — modifying and interpreting environmental traces — we propose a \emph{Stigmergic Multi-Agent Deep Reinforcement Learning (S-MADRL)} framework that leverages virtual pheromones to model local and social interactions, enabling decentralized emergent coordination without explicit communication.
To overcome the convergence and scalability limitations of existing algorithms such as MADQN, MADDPG, and MAPPO, we leverage curriculum learning, which decomposes complex tasks into progressively harder sub-problems. 
Simulation results show that our framework achieves the most effective coordination of up to eight agents, where robots self-organize into asymmetric workload distributions that reduce congestion and modulate group performance. This emergent behavior, analogous to strategies observed in nature, demonstrates a scalable solution for decentralized multi-agent coordination under crowded environments with communication constraints.

\end{abstract}

\section{Introduction}
\label{intro}

\begin{figure}[t!]
\begin{center}
\includegraphics[width=8.5cm, height=6cm]{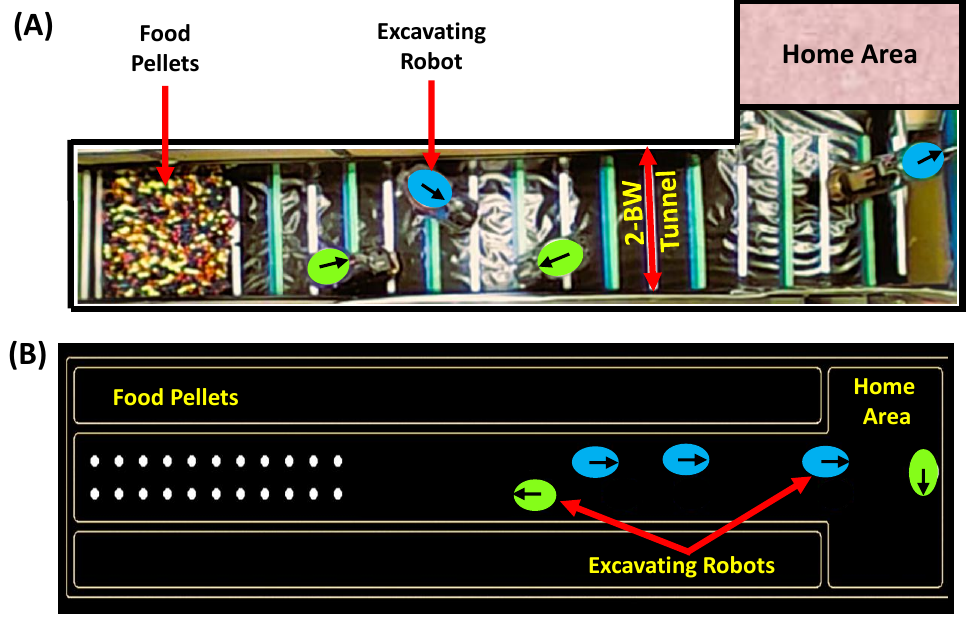}
\end{center}
\caption{Experimental setup of the multi-robot excavation task. \textbf{(A)} Real-world top view of the excavation arena, consisting of a pellet source, a narrow tunnel, excavating robots, and a home area. \textbf{(B)} Corresponding abstracted simulation model used for training and evaluation. The simplified representation preserves essential components in the real-world model, and enables scalable and varied experiment scenarios.} 
\label{fig:robot_sim_setup} 
\end{figure}

Social insect such as ants exhibit remarkable emergent coordination in tasks like nest construction or tunnel excavation. These behaviors remain robust regardless of colony size and rely on simple decentralized mechanisms such as local and social interactions like stigmergy, where agents leave traces in the environment that guide the actions of others ~\cite{doi_rsta_1198, Holland2002MultiagentS}. 
Such mechanisms allow large groups to operate effectively in constrained, shared environments where congestion and jamming would otherwise occur. In particular, stigmergy provides a bio-inspired form of indirect communication that enables efficient group performance in crowded and confined environments ~\cite{aguilar_collective_2018, Bruce2019, royal_interface_journal}.

\revised{
Translating these strategies into multi-robot systems, however, poses significant challenges ~\cite{aina2022toward, aina2025fault}. Achieving comparable levels of coordination and efficiency is hindered by the difficulty of accurately modeling the sensing and communication capabilities that underpin biological swarm interactions. In swarm robotics, researchers have often attempted to address this challenge through ad-hoc control laws or optimization methods. These typically rely on detailed modeling of the environment and robot interactions, frequently under assumptions such as full state observability that rarely hold in practice. An alternative approach involves designing simple behavioral rules or heuristics inspired by biological systems. While effective at producing complex group behaviors, such methods often require substantial re-modeling and adaptation when applied to new scenarios, thereby limiting their scalability and generalization.
}

\revised{
Deep reinforcement learning (DRL) offers the potential to discover complex behaviors from simple task descriptions. DRL has shown success in diverse domains including 
computer games \cite{mnih2013playing}, autonomous driving \cite{sallab2017deep}, and robotic manipulation \cite{Gu2016DeepRL}. However, extending DRL to swarm robotics or multi-agent systems presents significant challenges that hinder effective learning and stable convergence. First, agents face \textbf{partial observability}, as they lack access to global environmental information and rely solely on local observations. Second, the system is inherently \textbf{non-stationary}, with agents continuously adapting their behaviors or evolve their policies in response to others, destabilizing the learning process.  Finally, \textbf{non-scalability} arises as state and action spaces grow exponentially with the number of agents, exacerbating the ``curse of dimensionality."}

\revised{
In this work, we introduce the Stigmergic Multi-Agent Deep Reinforcement Learning (S-MADRL) framework, based on the conventional Deep Q-Network (DQN) algorithm \cite{Mnih2015HumanlevelCT}, to discover cooperative policies for collective pellet retrieval in crowded environments (see Figure \ref{fig:robot_sim_setup}). Our framework trains decentralized policies that operate without relying on global state information or joint actions of other agents, enabling effective coordination among homogeneous agents, reducing traffic jams, and enhancing overall pellet retrieval efficiency. The novelty of our approach lies in two key contributions. First, we implement the \textbf{stigmergic communication}, modeled as virtual pheromones, which locally encode traces of other agents' activities to enable indirect communication and foster cooperative behaviors. Second, we incorporate \textbf{curriculum learning} to address the issue of unstable learning in complex or crowded scenarios. By decomposing the problem into simpler sub-tasks, curriculum learning enables agents to effectively tackle increasingly complex scenarios, improving the stability and scalability of the learning process.}

\revised{
We evaluate our proposed framework on the collective pellet retrieval task, modeled using the OpenAI Gym platform \cite{brockman2016openai}. The results show that the SMADRL framework can effectively learn decentralized coordination behaviors for up to five agents in deterministic environments, and up to eight agents in stochastic environments. In contrast, state-of-the-art algorithms such as Multi-Agent Deep Deterministic Policy Gradient (MADDPG) 
fail to scale beyond two agents. Furthermore, the learned cooperative behaviors exhibit emergent strategies similar to those observed in biological systems ~\cite{aguilar_collective_2018}, emphasizing the potential of stigmergy-inspired approaches for scalable and robust multi-agent coordination in dynamic environments.}

\section{Literature Survey}
\label{literature}

The recent advancement of deep reinforcement learning (DRL) in tackling real-world tasks has established it as a powerful tool for solving complex problems characterized by high-dimensional information representation through trial-and-error methods ~\cite{Mnih2015HumanlevelCT}. Its extension to multi-agent systems (MAS) has gained substantial traction in recent years, enabling innovative collaborative and distributed solutions to challenges that traditional methods struggle to address ~\cite{vinyals2019grandmaster}. However, applying DRL to MAS presents significant challenges, including non-stationarity and non-convergence, primarily arising from the concurrent actions of agents and the partial observability of the state of the environment ~\cite{article_survey1}. To address these issues and enable convergence and scalability, prior research has explored several approaches, such as incorporating explicit communication channels between agents, training with joint action and full state information, and directly modeling the behavior of other agents ~\cite{article_survey1, Hernandez_Leal_2019, article_survey3}.

\revised{
Explicit communication in multi-agent systems often incurs significant computational costs, which do not scale well with an increasing number of agents. To address this limitation, recent works have explored implicit communication mechanisms or shared memory frameworks among agents ~\cite{article_survey1, lowe2020multiagent}. Stigmergy, in particular, has been investigated as a coherent and effective method of decentralized communication via local stimuli ~\cite{doi_rsta_1198}. Stigmergy ~\cite{stigmergy_article} enables agents to leave traces or signals in the environment, which guide the swarm towards desired emergent behaviors. This form of communication uses the environment as a shared external memory, where agents exchange useful information as cues. By interacting with these environmental cues, agents can infer the actions of neighbors, or approximate most recent activity history or state of their immediate environment ~\cite{NAYAK2020271}. Due to its scalability, robustness, low computational overhead, and adaptability ~\cite{585892}, stigmergy has been widely adopted in multi-agent coordination and control. Applications include solving several NP-hard problems, such as implicit spatial clustering of multi-agent systems without direct inter-agent communication ~\cite{article_owen} and coordination of unmanned vehicles ~\cite{digital_pheromone}.}


In multi-agent reinforcement learning (MARL) settings, adapting the concept of stigmergy offers the potential for better learning convergence and scalability, yet it remains under-explored. Conventional approaches to addressing the issues of non-stationarity and non-convergence often involve introducing direct communication between the agents, or training agents with joint actions and joint observations of all agents. However, these methods suffer from significant communication overhead, and the curse of dimensionality respectively ~\cite{article_survey1, lowe2020multiagent}. Consequently, such direct approaches face limitations in achieving scalability and convergence, particularly in congested or highly constrained environments. Recent studies have begun exploring the inherent advantages of stigmergy to improve learning in multi-agent settings \cite{9354492, pheromone_based_independent, nguyen2021scalable, Monekosso2001PheQAP, monekosso2001improved}.

Of particular relevance to our work is the Stigmergic Independent RL framework developed by Xu et al. \cite{9354492}. Their framework establishes an indirect communication bridge indirect communication bridge between agents via a stigmergy medium, represented as a digital pheromone map. This medium enables agents to communicate and observe environmental states through an explicit feedback loop. In contrast, our work adopts the concept of digital pheromones but employs a virtual map that stores localized information about different regions of the environment, avoiding direct information propagation. Furthermore, while Xu et al. applied their framework to UAV flight formation without addressing scalability, we focus on collective coordination in a highly constrained and crowded environment, providing a comparative scalability analysis with other popular MARL methods.

Similarly, Zhang et al \cite{pheromone_based_independent} developed the Pheromone Collaborative DQN framework to address the minefield navigation task. Their model uses a network structure where nodes emit pheromone information to attract nearby agents. Agents select ``attractors'' based on a random-proportional equation that depend on the attractor's signal strength. Unlike our model of digital pheromone model which directly influences agent decision-making through localized information, their approach relies on probabilistic attraction without analyzing scalability to larger numbers of agents. In contrast, we demonstrate our framework's effectiveness and scalability compared to other popular MADRL methods, specifically in dynamic and stochastic environments.

\section{Background}
\label{background}
\subsection{Deep Reinforcement Learning}
\label{background:1}
Deep reinforcement learning (DRL) typically models problems using a \textbf{Markov Decision Process (MDP)}. An MDP is defined by the state space $\mathcal{S}$, action space $\mathcal{A}$, transition function $p(s'|s,a)$, reward function $r(s,a)$, and the distribution of initial states $s_0 \sim p_0(s)$, where $s$ is the current state, $s'$ is the next state, and $a$ is the action. A widely-used method for solving MDPs is \textbf{Q-learning}, which learns the value $Q(s,a)$ of an action $a$ in a state, $s$, based on the \textbf{Temporal Difference (TD)} update rule below:
\begin{equation*}
    \quad Q(s,a) \leftarrow  Q(s,a) + \alpha(r + \gamma\max_{a'} Q(s',a') - Q(s,a)), \\
    \label{eqn:q_learning}
\end{equation*}

where $\alpha$ is the learning rate, $\gamma$ is the discount factor, $r$ is the reward, and $\max_{a'} Q(s',a')$ is the maximum future reward of the next state $s'$. Once $Q(s,a)$ is learned, the optimal policy $\pi(s)$ selects the action that maximizes the expected return: 
\begin{equation*}
    \qquad \qquad \qquad \pi(s) = \argmax_{a} Q(s,a).
\end{equation*}
 
\textbf{Deep Q-Network (DQN)} extends Q-learning by approximating the $Q-function$ using deep neural networks parameterized by $\theta$. Instead of directly updating individual $Q-values$, the network's parameters are updated by minimizing a differentiable loss function at each iteration $i$:
\begin{equation*}
 \qquad \qquad \mathcal{L(\theta_i)} \text{$ = \E_{s,a,r,s'}[(y_{i}^{DQN} - Q(s,a;\theta_i))^2], $}
    \label{eqn:dqn_loss}
\end{equation*}

where the target $y^{DQN}$ is given by:
\begin{equation*}
    \qquad \qquad \qquad y^{DQN} = r + \gamma\max_{a'}Q(s',a';\theta).
\end{equation*}

This approach enables the agent to generalize across high-dimensional state and action spaces, making DQN effective for a variety of complex tasks.

\subsection{Multi-Agent Deep Reinforcement Learning}
\label{background:2}

. 
Multi-agent Deep Reinforcement Learning (MADRL) extends the principles of DRL to address problems in multi-agent reinforcement learning (MARL). 
In these settings, the state transition function depends on the joint actions of all agents, introducing significant complexity. A naïve approach would involve learning a single policy for the joint actions and state spaces of all agents. However, this quickly becomes impractical due to the curse of dimensionality, where the state and action spaces grow exponentially with the number of agents. 

To address this, a common approach is to provide each agent with partial knowledge of the environment's state and to assign them individual policy learners. This reformulation is often modeled as a Partially Observable Markov Decision Process (POMDP). Under this framework, each agent operates based on a local observation rather than global state information, and learns separate decentralized policies tailored to their individual observations.
While this approach mitigates the scalability issue, it introduces a significant challenge: the non-stationarity problem. This arises because each agent continually updates its policy in response to the evolving behaviors of other agents, leading to a ``moving target" problem.

Several strategies have been proposed to address this issue: 1) team rewards which encourages cooperative behavior by providing agents with shared rewards that align their objectives, and 2) Direct Communication Channels allowing agents to exchange information to reduce uncertainty and improve coordination \cite{learning2communicate}. Despite these advancements, achieving scalable and stable learning in MADRL remains a key area of research, particularly in environments with high congestion. \\ 


\section{Multi-agent Clog Control}
\label{clog_control}

\revised{
The coordination of multi-agent systems has garnered significant research interest in recent years, with the complexity of the problem increasing exponentially as the number of agents rises ~\cite{samvelyan2019starcraft}. In this work, we address the challenge of multi-agent coordination within a collective excavation task, where a group of homogeneous robots is assigned the task of continuously retrieving pellets (or food items) in a narrow, confined 2D grid-world environment ~\cite{aina2022toward, aina2025fault}. The goal is to maximize pellet collection within a specified time frame.}
\revised{
The robots involved typically operate under constraints of limited sensing and communication capabilities, without centralized control or access to global state information regarding the actions and states of other agents. Consequently, much of the system's activity is spent resolving conflicts arising from overcrowding and competition for space.}
\revised{
We model this collective excavation problem as a decentralized Partially Observable Markov Decision Process (Dec-POMDP) ~\cite{Oliehoek_2008}, using independent Deep Q-Network (DQN) learners (IQL). Coordination between agents is achieved through specialized social interactions facilitated by stigmergy, allowing agents to infer the actions of others indirectly through the environment. To address more complex scenarios, we incorporate curriculum learning, which simplifies the learning process by training agents sequentially, focusing on one agent at a time. }

\subsection{Decentralized POMDP}
\label{clog_control:1}
In this section, we formalize the collective excavation task as a Decentralized Partially Observable Markov Decision Process (Dec-POMDP) to model the decision-making process of agents with partial and localized observations.


\textbf{State and Observation:} 

\revised{Consider a system of \emph{N} agents operating within the environment (as depicted in Figure ~\ref{fig:robot_sim_setup}). At each time step, each agent obtains a local observation consisting of two components: an egocentric, non-spatial discrete component and a geocentric, spatial component. The egocentric state encodes the internal mode of the agent, which can be one of the following states: \emph{Going-to-dig}, \emph{Digging}, \emph{Exit-digging}, \emph{Going-home}, \emph{Dumping}, \emph{Exit-home}, or \emph{Collision}. The geocentric components capture the agent’s \emph{position}, \emph{orientation}, \emph{previous action}, \emph{number of collisions}, \emph{distance to home}, and \emph{distance to pellet source}. These observations form an 8-dimensional vector space for each agent, which is further augmented with ``virtual pheromone'' information from neighboring cells within the agent’s restricted field-of-view. This design ensures that each agent has a compact, private observation space, enabling the technique to scale efficiently with a larger number of agents without necessitating changes to the network's input size.}

\textbf{Action Space:} 
\revised{We adopted a discrete space representation consisting of a set of five actions. At each time step, the agents can choose one possible action within the set namely: \emph{North}, \emph{South}, \emph{West}, \emph{East}, and \emph{Stop}. If the target location is occupied, the agent incurs a penalty and enters the collision state or mode, remaining stationary until it selects an unoccupied position.}

\textbf{Reward Function:} 

To facilitate rapid convergence, a reward shaping strategy is applied. The dense reward function is designed to encourage efficient learning and prompt the agents to achieve their objectives quickly while contributing to the collective mission. The reward function incorporates four distinct signals: 

\let\labelitemi\labelitemiii
\begin{itemize}
    \item \textbf{Distance Reward, $r_d(s,t)$}: The agent receives a reward of $+2.5$ for moving closer to the goal (either the pellet location or home area depending on the internal state) compared to the previous time step.
    \item \textbf{Collision Reward, $r_c(s,t)$}: A penalty of $-2.0$ if an unladen or unloaded agent collides with another agent. 
    \item \textbf{Pellet Pickup Reward, $r_p(s,t)$}: A reward of $+50$ is received when an agent successfully locates the pellet or food item. In global reward settings, this reward value is distributed equally among all agents, encouraging collective cooperation rather than individual competition. 
    To clarify, a global reward assigns the same reward value to all agents, regardless of their individual contributions, explicitly encouraging cooperative behavior. In contrast, a local reward allocates rewards based solely on each agent’s individual performance, which often promotes competitive rather than collaborative behavior. 
    \item \textbf{Successful Trip Reward, $r_s(s,t)$}: A reward of $+50$ is received when an agent successfully delivers a pellet to the home area. Again, in the global reward settings, this reward is distributed to all agents to further promote cooperative behavior.    
\end{itemize}
The complete reward function is defined as: 
\begin{equation*}
    r(s,t) = w_d r_d (s,t) + w_c r_c(s, t) + w_p r_p (s,t) + w_s r_s (s,t)
    \label{eqn:q_learning}
\end{equation*}

\revised{Where the weights $w_d, w_c, w_p,$ and $w_s$ are set to $0.2, 0.2, 0.2,$ and $0.4$ respectively for all the experiments.}

\begin{figure}[t!]
\begin{center}
\includegraphics[width=8.5cm]{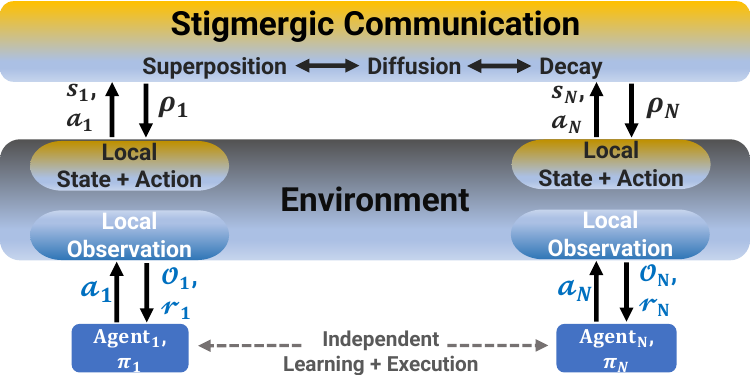}
\end{center}
\caption{Proposed scalable decentralized MADRL framework with stigmergic communication. Each agent $i$ receives a local observation $O_i$ and selects an action $a_i$ based on its policy $\pi_i$. The environment provides the resulting state $S_i$ and reward $r_i$, while agents leave and sense pheromone traces $\rho_i$ that diffuse and decay over time. This indirect communication channel encodes recent occupancy and agent activity, enabling decentralized coordination without explicit message passing. Learning and execution are fully independent for each agent, ensuring scalability to large team sizes.}
\label{fig:madrl_architecture} 
\end{figure}


\subsection{Digital Pheromone as Stigmergy}
\label{clog_control:2}

\begin{figure}[t!]
\begin{center}
\includegraphics[width=8.5cm]{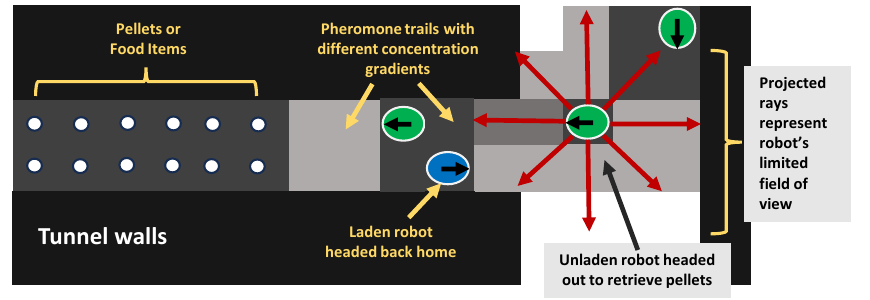}
\end{center}
\caption{Schematic of the digital pheromone map and agents’ restricted field of view. Agents deposit virtual pheromones while moving, generating spatial gradients (shown in different shades of gray) that diffuse and decay over time. Each agent perceives only a limited local region (red arrows), mimicking partial observability. The blue agent is laden with pellet and returning to the home area, while green agents are searching for pellets. Stigmergic communication provides environmental memory that supports implicit coordination among agents.}
\label{fig:pherom_map} 
\end{figure}

\revised{Figure~\ref{fig:madrl_architecture} illustrates our proposed  Stigmergic Multi-Agent Deep Reinforcement Learning (SMADRL) framework,  designed for scalable decentralized learning through the use of stigmergic communication. In this framework, we model stigmergy using a \emph{virtual map} overlaid on the environment, which serves as an asynchronous, indirect communication medium. As agents navigate the environment, they update this virtual map. 
A digital pheromone is essentially a virtual map that stores the distribution of \textbf{``activity signals''} across the environment. The map’s keys represent distinct coordinates or positions within the space, while the values encode specific state information about agents' \textbf{actions}. Each time an agent moves from one cell to another in the grid world (as depicted in Figure~\ref{fig:pherom_map}), the value of the corresponding map position is updated to reflect the agent’s state — such as its occupancy status, internal state, \emph{laden} or \emph{unladen} condition (represented as $1$ and $0$, respectively), and current action.
The occupancy status is initialized to a value of $\rho_0$, and decays over time based the function: $\rho(t+1) = (1 - \alpha)\,\rho(t) + \beta$, where $\alpha$ is the decay rate and $\beta$ is the reinforcement increment. This enables the pheromone concentration to reflect recent activity in the environment, encouraging indirect coordination. This digital pheromone mechanism allows agents to implicitly communicate information regarding their internal states and the environment, analogous to the pheromone traces left by social insects. Agents can then observe the encoded information within their neighboring cells, within their limited field of view, and incorporate it into their observation space during training and decision-making (see Figures~\ref{fig:madrl_architecture} and ~\ref{fig:pherom_map}). This approach enables agents to infer the actions and intentions of other agents, as well as gain \textbf{insights} into the state of the environment, facilitating coordinated behavior. }

\subsection{Curriculum Learning}
\label{clog_control:3}

\revised{Curriculum learning (CL) is an instructional strategy that sequences training experiences to enhance the learning process, either by accelerating learning or improving final performance. In reinforcement learning (RL) contexts, CL typically involves designing a progression of tasks, beginning with simple problems and gradually increasing their complexity. This progression aims to optimize both the asymptotic performance and training efficiency ~\cite{narvekar2020curriculumlearningreinforcementlearning}. In multi-agent reinforcement learning (MARL) settings, CL plays a crucial role in mitigating the challenges posed by non-stationarity, particularly when multiple agents concurrently adapt their policies in response to each other's changing behaviors. By limiting concurrent policy updates to a single agent at a time, while incrementally adding more agents, this approach can lead to substantial improvements in performance and convergence.}

\revised{In our framework, CL is employed to enhance learning convergence in more complex and congested scenarios, particularly when the number of agents exceeds four. The process begins with two agents learning decentralized policies for collective pellet retrieval in parallel. These two fully trained agents are referred to as the \emph{old agents}. Subsequently, additional agents are introduced sequentially, with each \emph{new agent} learning its policy from scratch while the policies of the old agents remain fixed. This strategy reduces the non-stationarity issues by minimizing the ``moving target" effect caused by multiple concurrent learners. Our experiments demonstrate that this method facilitates scalability, allowing us to effectively train systems with up to \textbf{eight agents}, especially when combined with the stigmergic communication technique.}

\section{Experimental Setup}
\label{experiment}
\revised{To evaluate the effectiveness of our proposed approach, we conducted simulation experiments using the setup illustrated in Figure~\ref{fig:robot_sim_setup}. The Markov Decision Process (MDP) was implemented through OpenAI's Gym interface ~\cite{brockman2016openai}, where the environment's step function takes actions as inputs and returns the subsequent states, rewards, and terminal flags. The environment is designed to accommodate a variable number of agents, which can be specified at the outset of each experiment, along with the length of the tunnel in terms of grid cells or the agents' body length (BL).}
\revised{The agents operate within a shared environment but employ separate policy learners and distinct experiences. At each time step, the agents act simultaneously and receive private observations, including either a global or local reward. To enhance training stability, we employed Double Q-learning ~\cite{van2016deep} and experience replay ~\cite{mnih2013playing}. The policy architecture consists of three hidden layers, each containing 128 neurons. An $\epsilon$-greedy action-selection strategy was used to ensure comprehensive exploration of the state and action spaces, with the exploration rate decaying from \emph{100\%} to \emph{2\%} over the first \emph{10\%} of the total training duration.}

\revised{We compared four distinct learning techniques, varying the number of agents from one to five:}


\let\labelitemi\labelitemiii
\begin{itemize}
    \item {Independent Q-Learning (IQL)}: This is our baseline method which utilizes independent DQNs with local rewards to train agents separately.
    \item {IQL with Global Reward (IQL+G)}: This method incorporates global rewards into the baseline method, to encourage cooperation among agents.
    \item {IQL with Global Reward and Stigmergy (IQL+GS)}: Building on \emph{(IQL+G)}, this method adds a virtual pheromone map, discussed in previous section, to mitigate non-staionarity and facilitate agent cooperation.
    \item {Stigmergy with Curriculum Learning (IQL+GSC)}: This approach combines the \emph{(IQL+GS)} method with curriculum learning, introducing one agent at a time during training to improve stability and convergence.
\end{itemize}

\revised{The training and testing were conducted on a desktop computer featuring an AMD Ryzen 12-core processor at 3.4 GHz, 32 GB RAM, and an Nvidia RTX 2070 GPU. Each experimental setup was run multiple times, with consistent learning curves observed across trials. During testing, we recorded the number of pellets excavated by individual agents and logged the cumulative total for all agents at each time step. The comparison of total pellets excavated across different learning methods is shown in Figure~\ref{fig:excav_pellets}.}

\revised{Due to the simulation's latency, training a single agent from scratch for a fully converged policy on the GPU takes approximately twelve hours. Training time scales linearly with the number of concurrent agents. However, curriculum learning (CL) significantly reduces training time by introducing new agents sequentially to a team of previously trained agents. Convergence typically begins after around $200$ episodes, each comprising $5000$ time steps, after which the environment is reset to its initial conditions. The learning rate was set to $10^{-4}$, with a discount factor of $0.99$. A batch size of 64 was found to be optimal, and the \emph{Adam optimizer} was used for agent training. }





\section{Simulation Results}
\label{results}




\begin{figure}[t!]
\begin{center}
\includegraphics[width=8.5cm]{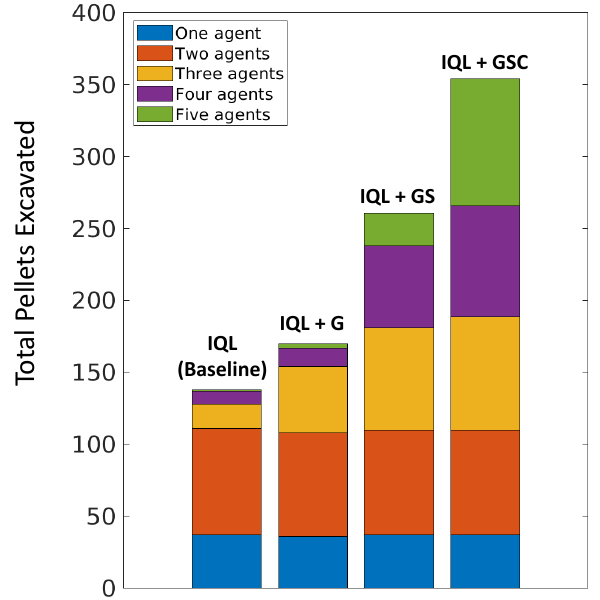}
\end{center}
\caption{Cumulative excavation results comparing four multi-agent deep reinforcement learning (MADRL) methods (IQL baseline, IQL+G, IQL+GS, and IQL+GSC) for teams of up to five agents. While the baseline (IQL) performs adequately for one and two agents, performance declines as team size increases. Incorporating stigmergy (IQL+GS) improves coordination for three and four agents, but struggles at five. Combining stigmergy with curriculum learning (IQL+GSC) achieves the highest excavation performance across all team sizes, demonstrating superior scalability and robustness in moderately congested environments.}
\label{fig:excav_pellets} 
\end{figure}


\begin{figure}[t!]
\begin{center}
\includegraphics[width=8.5cm]{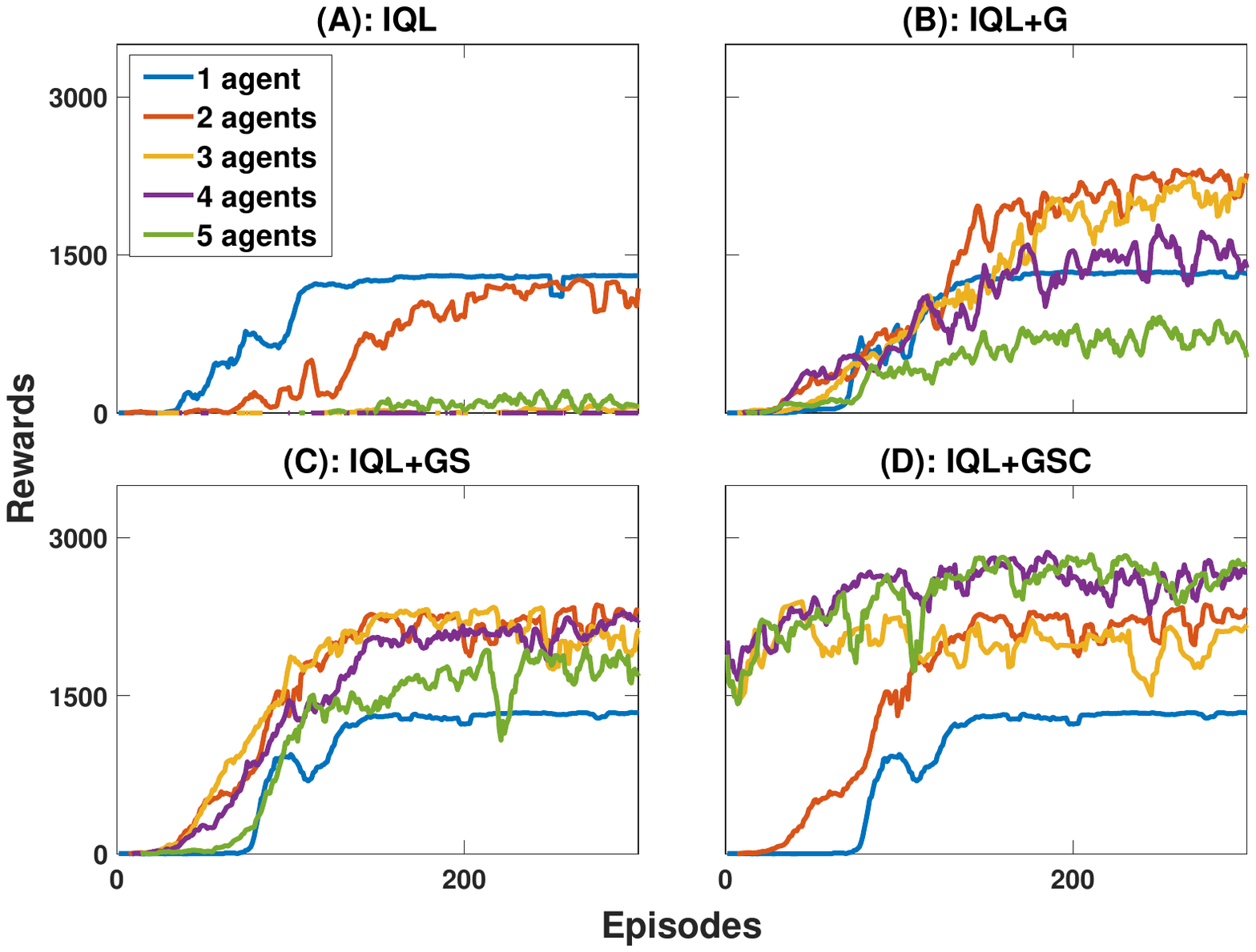}
\end{center}
\caption{Learning curve comparison of the four MADRL techniques (IQL, IQL+G, IQL+GS, and IQL+GSC) for one to five agents. Baseline methods (IQL, IQL+G) fail to converge beyond three agents. Stigmergic communication (IQL+GS) enhances learning stability, while curriculum learning (IQL+GSC) further accelerates convergence and achieves the highest rewards in four- and five-agent scenarios. 
}
\label{fig:learning_curves} 
\end{figure}


\begin{figure}[ht!]
\begin{center}
\includegraphics[width=8.5cm]{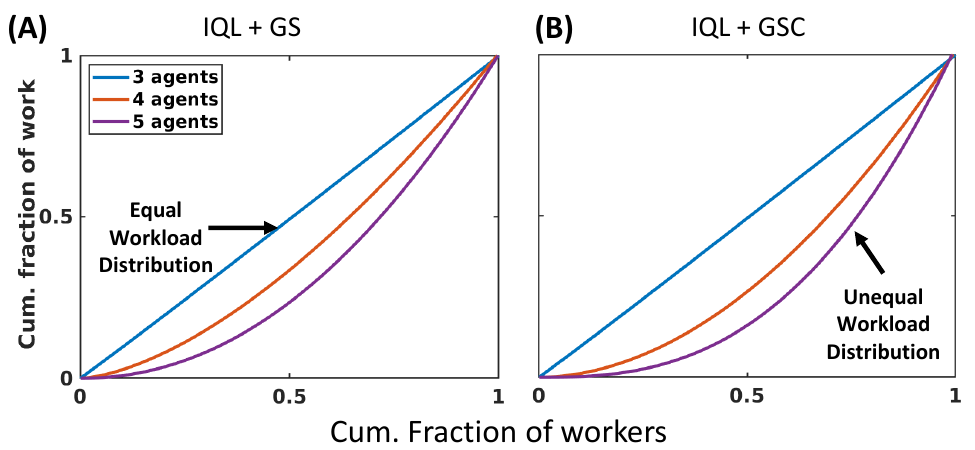}
\end{center}
\caption{Lorenz curve comparison of workload distribution under stigmergy-only learning (IQL+GS, panel \textbf{A}) and stigmergy with curriculum learning (IQL+GSC, panel \textbf{B}). For three agents, both methods result in equal workload distribution. For four and five agents, however, workload becomes increasingly unequal, especially under curriculum learning. 
}
\label{fig:lorenz_curves} 
\end{figure}


\revised{
Figure \ref{fig:excav_pellets} shows a comparison of total pellets excavated for each technique. For one and two agent scenarios, the baseline algorithm (IQL) performed well with both local and global rewards. However, as the number of agents increases, the performance drops drastically. At this stage, we introduce stigmergy to facilitate cooperation and to improve convergence. This is achieved by augmenting the observation space of each agent with information encoded in the virtual maps or ``pheromone trails". This technique improves performance for three and four team sizes, but not so well for five agent scenario. In this case, coordination becomes more challenging, and achieving successful convergence would require more powerful training technique. Our curriculum learning approach (IQL+GSC) addresses this by reducing the difficulty in four and five agent scenarios, training one agent at a time. }

Figure \ref{fig:learning_curves} shows the learning curve for the different MADRL techniques described above, for one to five agents in a shared environment. Clearly, stigmergy helps address convergence issues, since the baseline methods (IQL and IQL+G) could not handle three agents and higher. Curriculum learning improves learning convergence for $four$ and $five$ agent scenarios. Also, learning did not start from zero with the CL method, since the $old\ agents$ were using their learned policies.

Figure \ref{fig:lorenz_curves} shows the comparison of the workload distribution for the two best learning approaches using Lorenz curves \cite{Fellman2011}. Lorenz curves describe the workload distribution of a team by linking the cumulative share of number of agents to the cumulative share of total number of pellets or food items retrieved. This curve is convex by definition. An equal workload distribution appears as a straight line between (0,0) and (1,1). A divergence from this straight line indicates unequal workload distribution, where for example, half of the team retrieved less than half of the total number of pellets. According to the curves in Figure \ref{fig:lorenz_curves}, it is clear that for three agent scenario, all workers excavated equal amounts of pellets as indicated by the straight line in both figures. For four and five agents however, the curves diverge further from the straight line, with the curriculum learning technique (Figure \ref{fig:lorenz_curves}B) having the greater divergence or ``higher'' unequal workload distribution. This translates to: more agents should participate less as the team size increases. Thus, we can infer from the Lorenz curves and Figure \ref{fig:excav_pellets} that the best strategy to achieve the highest pellet retrieval successes is through asymmetric or unequal workload distribution when the agent density is above a threshold, in this case, four agents. This strategy greatly minimizes jamming and collision in the tunnel. Lorenz curves for one and two team sizes were not shown here because they all produce similar curves which are equal workload distributions.

\begin{figure}[t!]
\begin{center}
\includegraphics[width=8.3cm]{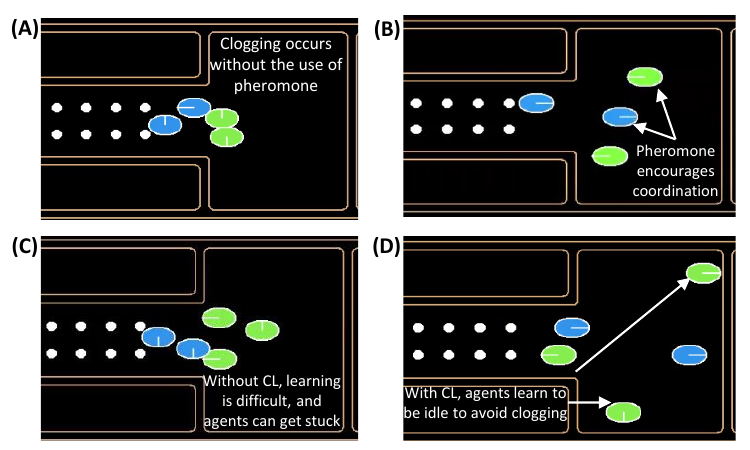}
\end{center}
\caption{Snapshots of various scenarios during testing. \textbf{(A)} Four agents often get stuck in the tunnel when trained without stigmergy. \textbf{(B)} Stigmergy prevents agents from getting stuck, which enables effective coordination. \textbf{(C)} Five agents with stigmergy can still get stuck due to the difficultly of the scenario. \textbf{(D)} Curriculum learning enables efficient learning in difficult scenarios, and the discovery of a bio-inspired strategy: unequal work distribution to avoid clogging.}
\label{fig:clog_vs_no_clog}
\end{figure}



Figure \ref{fig:clog_vs_no_clog} illustrates representative behaviors discovered by the different training methods. 
Figure \ref{fig:clog_vs_no_clog}A is a typical ``clogging'' or ``jamming" scenario we encounter during testing when agents are trained without incorporating stigmergy, even though global reward was used. The agents get stuck due to lack of cooperation. On the contrary, Figure \ref{fig:clog_vs_no_clog}B shows the coordinated behavior that emerges when virtual pheromone is employed. The agents learned to carefully avoid clogging by spacing out when outside the tunnel and yielding to each other when inside the tunnel. This is effective to avoid getting stuck in three to four agent scenarios. In five agent scenario, the agents can still get stuck even with virtual pheromones (Figure \ref{fig:clog_vs_no_clog}C). This is due to the complexity of the problem: high non-stationarity. Incorporating curriculum learning resolves this: some agents strategically remain idle, while others complete retrieval tasks (Figure \ref{fig:clog_vs_no_clog}D). 
The snapshot shows that two agents are ``idle" or ``resting" by not participating in the pellet retrieval task 
(\emph{follow \href{https://youtube.com/playlist?list=PLUk3lwDzxP0zwoo8N-7Nn7X52gtzvtrq5}{link} for video demonstration}). This emergent division of labor mirrors biological findings that idleness and asymmetric workload distribution are effective congestion-control strategies in crowded environments~\cite{aguilar_collective_2018}. 


\section{Discussion}
\label{discussion}

\begin{table*}[t]
\centering
\scriptsize
\caption{Quantitative comparison of excavation performance across multiple multi-agent deep reinforcement learning algorithms, measured as the total number of successful trips (pellets delivered per episode) for teams of 1–8 agents. 
Our proposed S-MADRL framework consistently scales up to 8 agents, maintaining robust performance under severe congestion.}
\label{tab:comparison}
\setlength{\tabcolsep}{4pt}
\begin{tabular}{lcccccccc}
\toprule
\textbf{Method} & \textbf{1 Agent} & \textbf{2 Agents} & \textbf{3 Agents} & \textbf{4 Agents} & \textbf{5 Agents} & \textbf{6 Agents} & \textbf{7 Agents} & \textbf{8 Agents} \\
\midrule

I-DQN        & 43 & 63 & 99 & 73 & 79 & 43 & 32 & 35 \\
I-A2C        & 53 & 69 & 79 & 49 & 22 & 8 & 6 & 3 \\
MA-DQN       & 32 & 15 & 11 & 8 & 1 & 2 & 0 & 0 \\
MA-A2C       & 52 & 36 & 5 & 0 & 0 & 0 & 0 & 0 \\
MADDPG       & 33 & 33 & 17 & 28 & 3 & 3 & 1 & 1 \\
MAPPO       & 43 & 99 & 92 & 57 & 21 & 22 & 3 & 2 \\
S-MADRL      & 46 & 87 & 102 & 115 & 114  & 105 & 109 & 110 \\

\bottomrule
\end{tabular}
\end{table*}

\begin{figure*}[t] 
    \centering
    \includegraphics[width=15cm]{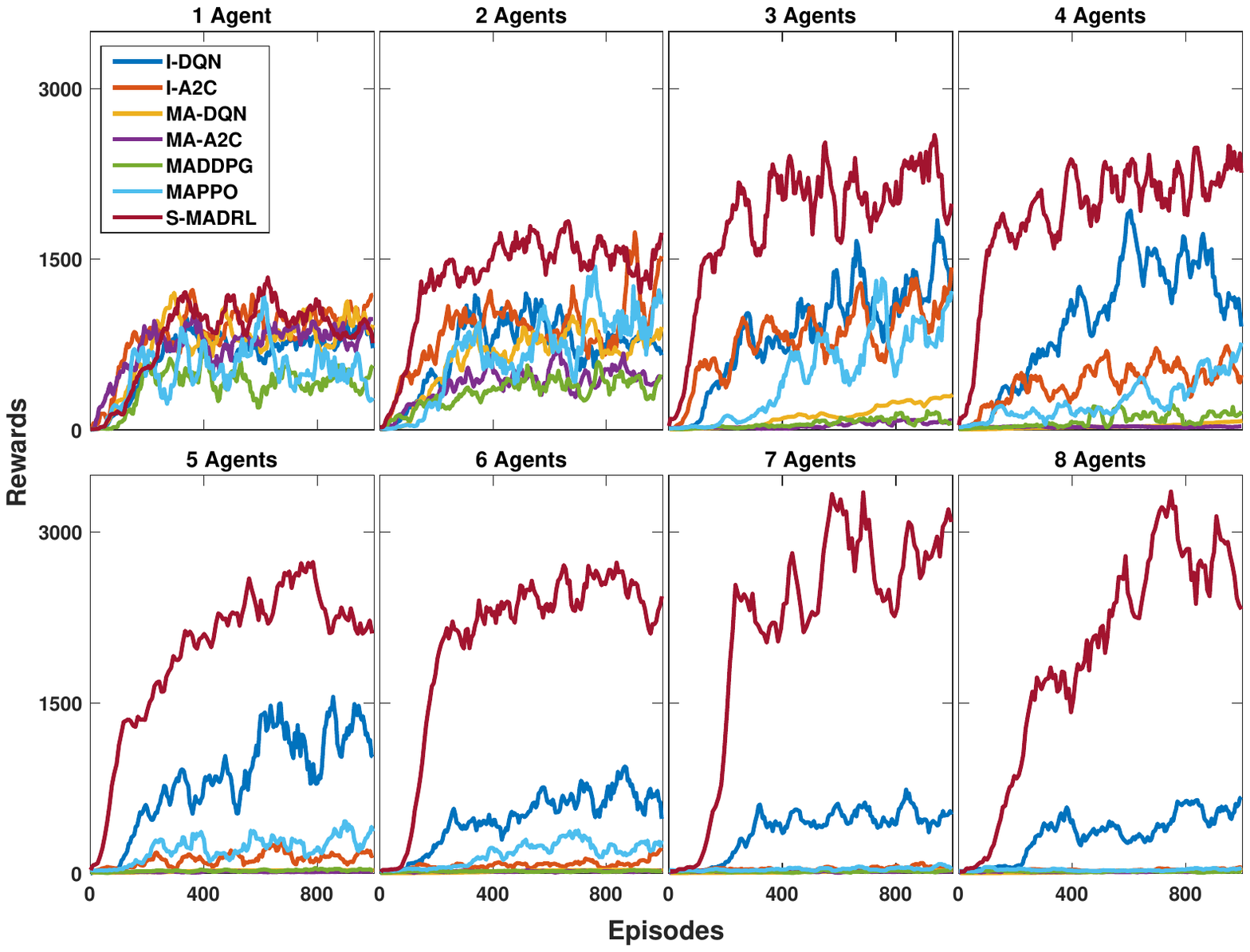} %
    \caption{Learning curve comparison of our proposed S-MADRL framework against state-of-the-art multi-agent deep RL baseline algorithms for team sizes from 1 to 8 agents. Each subplot shows cumulative rewards over training episodes. Baseline methods achieve moderate success with 1–3 agents but collapse beyond 4 agents due to high dimensional state and action spaces. In contrast, S-MADRL demonstrates stable convergence and sustained high rewards even with 7–8 agents, confirming its superior scalability and coordination capabilities in crowded environments.}
    \label{fig:learning_curve_cmp}
\end{figure*}

Scaling multi-agent deep reinforcement learning to dense and dynamic environments remains a central challenge due to the compounded non-stationarity introduced by many concurrent learners interacting in a shared evolving environment. As shown in Table~\ref{tab:comparison} and Figure~\ref{fig:learning_curve_cmp}, conventional approaches such as I-DQN, I-A2C, MA-DQN, MA-A2C, MADDPG, and MAPPO degrade rapidly as the number of agents increases, often failing to converge beyond 3–4 agents. By contrast, our proposed S-MADRL framework maintains stability and scales effectively up to eight agents, representing a significant improvement over state-of-the-art baselines.

A key driver of this scalability is the incorporation of tunnel density information ($\rho = n/L$, where $n$ is the number of agents in the tunnel and $L$ the tunnel length) into the pheromone map. This density signal enables agents to infer congestion levels and adaptively switch between active mode (entering the tunnel) and passive mode (remaining idle at home). This emergent behavior mirrors strategies observed in social insects, where temporary idleness prevents clogging. By leveraging environmental memory via stigmergy, agents coordinate indirectly, mitigating non-stationarity without explicit communication or centralized critics.

From qualitative analysis, we observed two recurring optimal policy strategies (\emph{follow \href{https://youtube.com/playlist?list=PLUk3lwDzxP0zwoo8N-7Nn7X52gtzvtrq5}{link} for video demonstration}):

\let\labelitemi\labelitemiii
\begin{itemize}
    \item {One-at-a-Time (OAT) strategy~\cite{aina2022toward, royal_interface_journal}}: This is more frequently observed in joint-action learners (JALs) such as MA-DQN, MADDPG, and MAPPO, where only two agents operate at a time while others wait. While this reduces tunnel congestion, it creates bottlenecks at the home area as team size grows, limiting group performance and scalability.
    \item {Bucket-Brigade (BB) strategy~\cite{anderson2002task}}: This is common in independent learners (I-DQN, I-A2C) and strongly reinforced by our S-MADRL. Here, agents self-organize into a bi-directional flow pattern: one lane for unladen robots entering the tunnel and the other for laden robots returning home. This continuous flow minimizes interference and scales efficiently with team size (Table~\ref{tab:comparison} and Figure~\ref{fig:learning_curve_cmp}), underpinning the superior performance of our S-MADRL framework.
\end{itemize}

Our results suggest that indirect communication through stigmergic traces is not only biologically plausible but also computationally advantageous: it provides a spatio-temporal signal that agents can exploit for traffic coordination, avoiding the exponential growth of state-action spaces typical in explicit communication or centralized learning settings.

Beyond simulation, S-MADRL offers direct applications for real-world robotic swarms operating in highly constrained domains such as mining excavations or search-and-rescue missions. Training can be carried out in simulation to learn robust policies, which can then be transferred to physical robots provided that the extracted features in observation remain consistent, or are finetuned during deployment. In practice, virtual pheromone maps could be approximated through several mechanisms: for instance, via synchronized distributed servers aggregating and diffusing local agent observations; 
or through physical environmental cues such as dynamically updated QR-code projections or RFID markers that allow agents to project and sense shared state information. 

Unlike many existing MADRL approaches, our method does not require direct inter-agent communication or sharing of policy parameters, or centralized training and execution, which typically suffer from scalability bottlenecks due to exponential state–action growth. Instead, S-MADRL remains fully decentralized in both training and execution, making it especially suitable for hardware deployment in environments with unreliable or degraded communications. Nevertheless, our approach assumes accurate sensing of pheromone intensities which may be affected by noise in real-world systems, and the restriction to homogeneous agents. Future work will extend S-MADRL to heterogeneous roles (e.g., specialized excavators vs. transporters) and refine curriculum learning to reduce its linear dependence on team size.

\section{Conclusion and Future Work}
\label{conclusions}

This work advances the field of multi-agent deep reinforcement learning by demonstrating how stigmergic communication and curriculum learning together enable robust scalability in environments that are otherwise dominated by congestion and non-stationarity. Simulation results demonstrate that S-MADRL not only improves excavation efficiency but also yields emergent, biologically inspired strategies such as selective idleness and asymmetric workload distribution, which reduce congestion in confined tunnels. Benchmark comparisons confirm that our framework scales reliably up to eight agents, outperforming state-of-the-art baselines.
The emergence of biologically inspired behaviors such as asymmetric workload distribution and selective idleness, underscores the value of indirect communication as a robust and scalable multi-agent coordination mechanism.
Looking forward, we believe that our approach holds promise for addressing other multi-agent coordination problems, in particular those that required decentralized and scalable control framework, in both static and dynamic contexts.

\begin{acknowledgement}
We gratefully acknowledge Professor Daniel I. Goldman of the Georgia Institute of Technology for his invaluable guidance and support throughout the course of this research. His insights and feedback were instrumental in shaping both the conceptual framework and the successful completion of this manuscript.

\end{acknowledgement}

\newpage

\bibliographystyle{unsrt} 
\bibliography{bibliography}

\begin{thebibliography}{10}

\bibitem{doi_rsta_1198}
J.~Skår, P.~V. Coveney, Guy Theraulaz, Jacques Gautrais, Scott Camazine, and Jean-Louis Deneubourg.
\newblock The formation of spatial patterns in social insects: from simple behaviours to complex structures.
\newblock {\em Philosophical Transactions of the Royal Society of London. Series A: Mathematical, Physical and Engineering Sciences}, 361(1807):1263--1282, 2003.

\bibitem{Holland2002MultiagentS}
OE~Holland.
\newblock Multiagent systems: Lessons from social insects and collective robotics.
\newblock In {\em Adaptation, Coevolution and Learning in Multiagent Systems: Papers from the 1996 AAAI Spring Symposium}, pages 57--62, 1996.

\bibitem{aguilar_collective_2018}
J.~Aguilar, D.~Monaenkova, V.~Linevich, W.~Savoie, B.~Dutta, H.-S. Kuan, M.~D. Betterton, M.~a.~D. Goodisman, and D.~I. Goldman.
\newblock Collective clog control: {{Optimizing}} traffic flow in confined biological and robophysical excavation.
\newblock {\em Science}, 361(6403):672--677, August 2018.

\bibitem{Bruce2019}
A.~I. Bruce, A.~P{\'{e}}rez-Escudero, T.~J. Czaczkes, and M.~Burd.
\newblock {The digging dynamics of ant tunnels: movement, encounters, and nest space}.
\newblock {\em Insectes Sociaux}, 66(1):119--127, feb 2019.

\bibitem{royal_interface_journal}
R.~Avinery, K.~O. Aina, C.~J. Dyson, H.~Kuan, M.~D. Betterton, M.~A.~D. Goodisman, and D.~I. Goldman.
\newblock Agitated ants: regulation and self-organization of incipient nest excavation via collisional cues.
\newblock {\em Journal of the Royal Society Interface}, 20, 2023.

\bibitem{aina2022toward}
Kehinde~O Aina, Ram Avinery, Hui-Shun Kuan, Meredith~D Betterton, Michael~AD Goodisman, and Daniel~I Goldman.
\newblock Toward task capable active matter: learning to avoid clogging in confined collectives via collisions.
\newblock {\em Frontiers in Physics}, 10:735667, 2022.

\bibitem{aina2025fault}
Kehinde~O Aina, Hosain Bagheri, and Daniel~I Goldman.
\newblock Fault-tolerant multi-robot coordination with limited sensing within confined environments.
\newblock {\em arXiv preprint arXiv:2505.15036}, 2025.

\bibitem{mnih2013playing}
Volodymyr Mnih, Koray Kavukcuoglu, David Silver, Alex Graves, Ioannis Antonoglou, Daan Wierstra, and Martin Riedmiller.
\newblock Playing atari with deep reinforcement learning.
\newblock {\em arXiv preprint arXiv:1312.5602}, 2013.

\bibitem{sallab2017deep}
Ahmad~EL Sallab, Mohammed Abdou, Etienne Perot, and Senthil Yogamani.
\newblock Deep reinforcement learning framework for autonomous driving.
\newblock {\em Electronic Imaging}, 2017(19):70--76, 2017.

\bibitem{Gu2016DeepRL}
Shixiang~Shane Gu, Ethan Holly, Timothy~P. Lillicrap, and Sergey Levine.
\newblock Deep reinforcement learning for robotic manipulation.
\newblock {\em ArXiv}, abs/1610.00633, 2016.

\bibitem{Mnih2015HumanlevelCT}
V.~Mnih, K.~Kavukcuoglu, D.~Silver, Andrei~A. Rusu, J.~Veness, Marc~G. Bellemare, A.~Graves, Martin~A. Riedmiller, A.~Fidjeland, Georg Ostrovski, Stig Petersen, Charlie Beattie, A.~Sadik, Ioannis Antonoglou, Helen King, D.~Kumaran, Daan Wierstra, S.~Legg, and D.~Hassabis.
\newblock Human-level control through deep reinforcement learning.
\newblock {\em Nature}, 518:529--533, 2015.

\bibitem{brockman2016openai}
Greg Brockman, Vicki Cheung, Ludwig Pettersson, Jonas Schneider, John Schulman, Jie Tang, and Wojciech Zaremba.
\newblock Openai gym, 2016.

\bibitem{vinyals2019grandmaster}
Oriol Vinyals, Igor Babuschkin, Wojciech~M Czarnecki, Micha{\"e}l Mathieu, Andrew Dudzik, Junyoung Chung, David~H Choi, Richard Powell, Timo Ewalds, Petko Georgiev, et~al.
\newblock Grandmaster level in starcraft ii using multi-agent reinforcement learning.
\newblock {\em Nature}, 575(7782):350--354, 2019.

\bibitem{article_survey1}
Lucian Busoniu, Robert Babuska, and Bart De~Schutter.
\newblock A comprehensive survey of multiagent reinforcement learning.
\newblock {\em Systems, Man, and Cybernetics, Part C: Applications and Reviews, IEEE Transactions on}, 38:156 -- 172, 04 2008.

\bibitem{Hernandez_Leal_2019}
Pablo Hernandez-Leal, Bilal Kartal, and Matthew~E. Taylor.
\newblock A survey and critique of multiagent deep reinforcement learning.
\newblock {\em Autonomous Agents and Multi-Agent Systems}, 33(6):750–797, Oct 2019.

\bibitem{article_survey3}
Sven Gronauer and Klaus Dieopold.
\newblock Multi-agent deep reinforcement learning: a survey.
\newblock {\em Artificial Intelligence Review}, pages 1--49, 04 2021.

\bibitem{lowe2020multiagent}
Ryan Lowe, Yi~Wu, Aviv Tamar, Jean Harb, Pieter Abbeel, and Igor Mordatch.
\newblock Multi-agent actor-critic for mixed cooperative-competitive environments.
\newblock In {\em Proceedings of the 31st International Conference on Neural Information Processing Systems}, NIPS'17, page 6382–6393, Red Hook, NY, USA, 2017. Curran Associates Inc.

\bibitem{stigmergy_article}
Guy Theraulaz and Eric Bonabeau.
\newblock A brief history of stigmergy.
\newblock {\em Artificial life}, 5:97--116, 02 1999.

\bibitem{NAYAK2020271}
Sukanta Nayak.
\newblock Chapter ten - nature-inspired optimization.
\newblock In Sukanta Nayak, editor, {\em Fundamentals of Optimization Techniques with Algorithms}, pages 271--296. Academic Press, 2020.

\bibitem{585892}
M.~Dorigo and L.M. Gambardella.
\newblock Ant colony system: a cooperative learning approach to the traveling salesman problem.
\newblock {\em IEEE Transactions on Evolutionary Computation}, 1(1):53--66, 1997.

\bibitem{article_owen}
Owen Holland and Chris Melhuish.
\newblock Stimergy, self-organization, and sorting in collective robotics.
\newblock {\em Artificial Life}, 5:173--202, 04 1999.

\bibitem{digital_pheromone}
H.~Van Dyke~Parunak, Sven~A. Brueckner, and John Sauter.
\newblock Digital pheromones for coordination of unmanned vehicles.
\newblock In Danny Weyns, H.~Van Dyke~Parunak, and Fabien Michel, editors, {\em Environments for Multi-Agent Systems}, pages 246--263, Berlin, Heidelberg, 2005. Springer Berlin Heidelberg.

\bibitem{9354492}
Xing Xu, Rongpeng Li, Zhifeng Zhao, and Honggang Zhang.
\newblock Stigmergic independent reinforcement learning for multiagent collaboration.
\newblock {\em IEEE Transactions on Neural Networks and Learning Systems}, pages 1--15, 2021.

\bibitem{pheromone_based_independent}
Kaige Zhang, Yaqing Hou, Hua Yu, Wenxuan Zhu, Liang Feng, and Qiang Zhang.
\newblock Pheromone based independent reinforcement learning for multiagent navigation.
\newblock In Haijun Zhang, Zhi Yang, Zhao Zhang, Zhou Wu, and Tianyong Hao, editors, {\em Neural Computing for Advanced Applications}, pages 44--58, Singapore, 2021. Springer Singapore.

\bibitem{nguyen2021scalable}
Austin~Anhkhoi Nguyen.
\newblock Scalable, decentralized multi-agent reinforcement learning methods inspired by stigmergy and ant colonies.
\newblock {\em arXiv preprint arXiv:2105.03546}, 2021.

\bibitem{Monekosso2001PheQAP}
Dorothy~Ndedi Monekosso and Paolo Remagnino.
\newblock Phe-q: A pheromone based q-learning.
\newblock In {\em Australian Joint Conference on Artificial Intelligence}, 2001.

\bibitem{monekosso2001improved}
Ndedi Monekosso, Paolo Remagnino, and Adam Szarowicz.
\newblock An improved q-learning algorithm using synthetic pheromones.
\newblock In {\em International Workshop of Central and Eastern Europe on Multi-Agent Systems}, pages 197--206. Springer, 2001.

\bibitem{learning2communicate}
Jakob~N. Foerster, Yannis~M. Assael, Nando de~Freitas, and Shimon Whiteson.
\newblock Learning to communicate with deep multi-agent reinforcement learning.
\newblock In {\em Proceedings of the 30th International Conference on Neural Information Processing Systems}, NIPS'16, page 2145–2153, Red Hook, NY, USA, 2016. Curran Associates Inc.

\bibitem{samvelyan2019starcraft}
Mikayel Samvelyan, Tabish Rashid, Christian~Schroeder De~Witt, Gregory Farquhar, Nantas Nardelli, Tim~GJ Rudner, Chia-Man Hung, Philip~HS Torr, Jakob Foerster, and Shimon Whiteson.
\newblock The starcraft multi-agent challenge.
\newblock {\em arXiv preprint arXiv:1902.04043}, 2019.

\bibitem{Oliehoek_2008}
F.~A. Oliehoek, M.~T.~J. Spaan, and N.~Vlassis.
\newblock Optimal and approximate q-value functions for decentralized pomdps.
\newblock {\em Journal of Artificial Intelligence Research}, 32:289–353, May 2008.

\bibitem{narvekar2020curriculumlearningreinforcementlearning}
Sanmit Narvekar, Bei Peng, Matteo Leonetti, Jivko Sinapov, Matthew~E. Taylor, and Peter Stone.
\newblock Curriculum learning for reinforcement learning domains: A framework and survey, 2020.

\bibitem{van2016deep}
Hado Van~Hasselt, Arthur Guez, and David Silver.
\newblock Deep reinforcement learning with double q-learning.
\newblock In {\em Proceedings of the AAAI conference on artificial intelligence}, volume~30, 2016.

\bibitem{Fellman2011}
Johan Fellman.
\newblock {\em Lorenz Curve}, pages 760--762.
\newblock Springer Berlin Heidelberg, Berlin, Heidelberg, 2011.

\bibitem{anderson2002task}
Carl Anderson, Jacobus~J Boomsma, and John~J Bartholdi, III.
\newblock Task partitioning in insect societies: bucket brigades.
\newblock {\em Insectes Sociaux}, 49(2):171--180, 2002.

\end{thebibliography}


\end{document}